\documentclass[letterpaper, 10 pt, journal, twoside]{IEEEtran}
\usepackage{cite}

\ifCLASSINFOpdf
\else
\fi
\usepackage{amsmath,amsfonts}
\usepackage{algorithmic}
\usepackage{array}
\usepackage[caption=false,font=normalsize,labelfont=sf,textfont=sf]{subfig}
\usepackage{textcomp}
\usepackage{stfloats}
\usepackage{url}
\usepackage{verbatim}
\usepackage{graphicx}
\usepackage{caption}
\def\BibTeX{{\rm B\kern-.05em{\sc i\kern-.025em b}\kern-.08em
    T\kern-.1667em\lower.7ex\hbox{E}\kern-.125emX}}
\usepackage{balance}
\usepackage{epsfig}
\usepackage{amsmath}
\usepackage{amssymb}
\usepackage{booktabs}
\usepackage{multirow}
\usepackage{multicol}
\usepackage[ruled,vlined]{algorithm2e}
\usepackage{pifont}
\usepackage[dvipsnames, svgnames, x11names]{xcolor}

\usepackage{colortbl}       
\usepackage{pifont}
\usepackage{graphicx} 
\IEEEoverridecommandlockouts   
\IEEEoverridecommandlockouts                              

\newcommand{\methodname}{{TinyVLA}\xspace}

\definecolor{convcolor}{HTML}{412F8A}
\definecolor{resnetcolor}{HTML}{8DA0CB}
\definecolor{vitcolor}{HTML}{fc8e62}

\newcommand{\convcolor}[1]{\textcolor{convcolor}{#1}}
\newcommand{\vitcolor}[1]{\textcolor{vitcolor}{#1}}

\newcommand{\vb}{\vitcolor{$\mathbf{\circ}$\,}}
\newcommand{\cb}{\convcolor{$\bullet$\,}}

\usepackage{booktabs}                                   
\usepackage{multirow}                                   
\usepackage{makecell}                                   
\usepackage{tablefootnote}                              
\usepackage[symbol]{footmisc}                           
\usepackage{amsmath,amssymb}                            
\usepackage{xcolor}                                     
\usepackage{enumitem}                                   
\usepackage{subcaption}                                 
\usepackage{stfloats}                                   
\usepackage[misc]{ifsym}     
\usepackage{hyperref}   

\newcommand{\eat}[1]{}                                  

\begin{document}
%
\title{TinyVLA: Towards Fast, Data-Efficient Vision-Language-Action Models for Robotic Manipulation}

%
%
%

\author{Junjie Wen$^{1,3}$,~Yichen Zhu$^{2}$,~\IEEEmembership{Member,~IEEE},~Jinming Li$^{3,6}$,~Minjie Zhu$^{1,3}$,~Zhibin Tang$^{2}$,~Kun Wu$^{4}$,~Zhiyuan Xu$^{5}$, \\ Ning Liu$^{5}$, Ran Cheng$^{2}$, Chaomin Shen$^{1}$, Yaxin Peng$^{6}$, Feifei Feng$^{2}$, and Jian~Tang$^{5}$,~\IEEEmembership{Fellow,~IEEE}\\
\url{https://tiny-vla.github.io/}
\vspace{-0.6cm}
\thanks{Manuscript received September 27, 2024; Revised December 28, 2024; Accepted February 6, 2025.}
\thanks{This paper was recommended for publication by
Editor Markus Vincze upon evaluation of the Associate Editor and Reviewers’
comments.
This work is supported by the Sci-Tech Innovation Initiative by the Science and Technology Commission of Shanghai Municipality (24ZR1419000),  and the National Science Foundation of China (12471501).} 
\thanks{$^{1}$Junjie Wen, Minjie Zhu, and Chaomin Shen are with East China Normal University, Shanghai 200042, China. 
        {\tt\footnotesize \{jjwen, mjzhu\}@stu.ecnu.edu.cn, cmshen@cs.ecnu.edu.cn}}%
\thanks{$^{2}$Yichen Zhu, Ran Cheng, Zhibin Tang, and Feifei Feng are with Midea Group, AI Lab, Shanghai 201700, China. 
        {\tt\footnotesize \{zhuyc25, tangzb, ningliu22, chengran, feifei.feng\}@midea.com}}%
\thanks{$^{3}$Junjie Wen, Minjie Zhu, and Jinming Li are interned at Midea Group, AI Lab, Shanghai 201700, China.}%
\thanks{$^{4}$Kun Wu is with Syracuse University, New York 13244, USA. 
        {\tt\footnotesize kwu102@syr.edu}}%
\thanks{$^{5}$Zhiyuan Xu, Ning Liu, and Jian Tang are with Beijing Innovation Center of Humanoid Robotics, Beijing 102676, China. 
        {\tt\footnotesize \{eric.xu, neil.liu, jian.tang\}@x - humanoid.com}}%
\thanks{$^{6}$Jinming Li and Yaxin Peng are with Shanghai University, Shanghai 201900, China. 
        {\tt\footnotesize \{ljm2022, yaxin.peng\}@shu.edu.cn}}%
\thanks{Junjie Wen and Yichen Zhu are co-first authors. Yichen Zhu and Chaomin Shen are the corresponding authors.}
\thanks{Digital Object Identifier (DOI): see top of this page.}
}
%
%

\markboth{IEEE ROBOTICS AND AUTOMATION LETTERS. PREPRINT VERSION. ACCEPTED
FEB 2025}
{WEN \MakeLowercase{\textit{et al.}}: TinyVLA: Towards Fast, Data-Efficient Vision-Language-Action Models for Robotic Manipulation} 

%



\maketitle

\begin{abstract}
Vision-Language-Action (VLA) models have shown remarkable potential in visuomotor control and instruction comprehension through end-to-end learning processes. However, current VLA models face significant challenges: they are slow during inference and require extensive pre-training on large amounts of robotic data, making real-world deployment difficult. In this paper, we introduce a new family of compact vision-language-action models, called TinyVLA, which offers two key advantages over existing VLA models: (1) faster inference speeds, and (2) improved data efficiency, eliminating the need for pre-training stage. Our framework incorporates two essential components to build TinyVLA: (1) initializing the policy backbone with robust, high-speed multimodal models, and (2) integrating a diffusion policy decoder during fine-tuning to enable precise robot actions. We conducted extensive evaluations of TinyVLA in both simulation and on real robots, demonstrating that our approach significantly outperforms the state-of-the-art VLA model, OpenVLA, in terms of speed and data efficiency, while delivering comparable or superior performance. Additionally, TinyVLA exhibits strong generalization capabilities across various dimensions, including language instructions, novel objects, unseen positions, changes in object appearance, background variations, and environmental shifts, often matching or exceeding the performance of OpenVLA. We believe that \methodname offers an interesting perspective on utilizing pre-trained multimodal models for policy learning.
\end{abstract}


\section{Introduction}
\begin{figure}[t]
    \centering
    \includegraphics[width=0.48\textwidth]{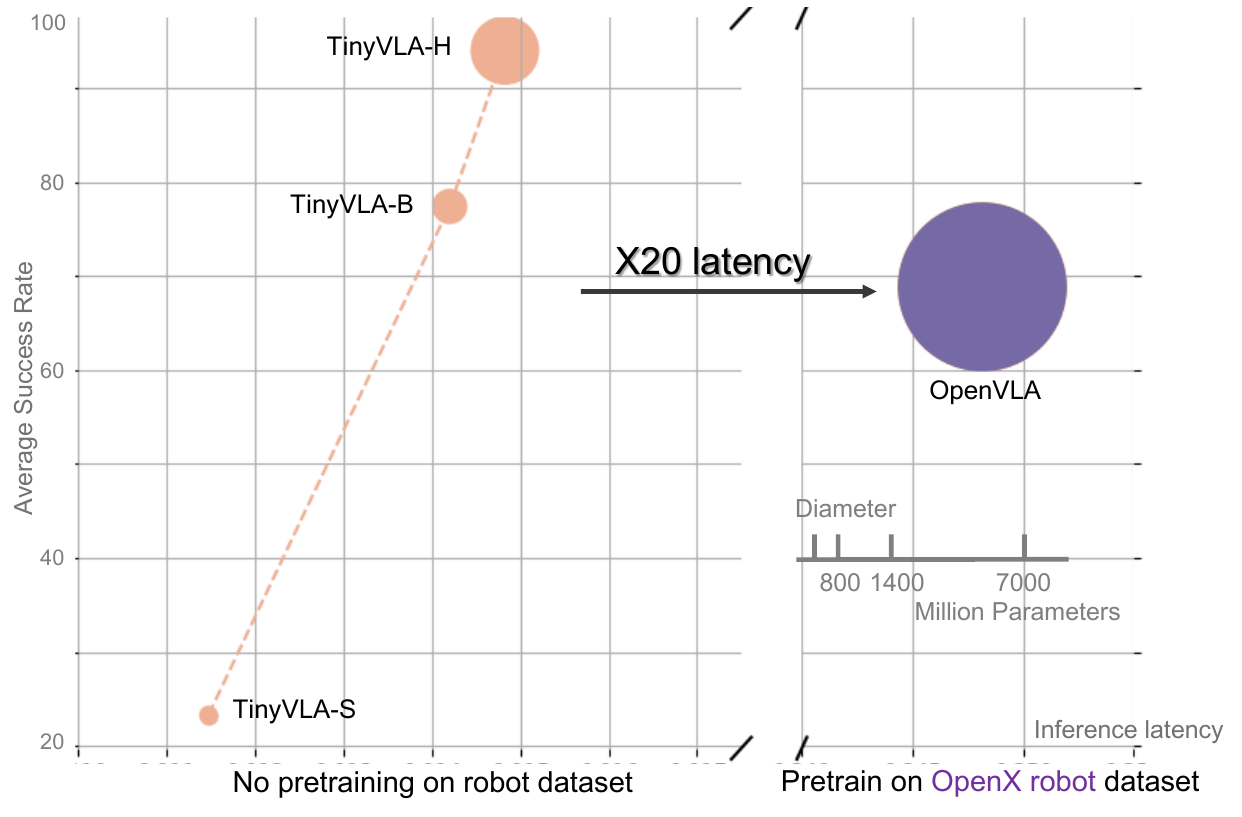}
    \caption{\textbf{Inference latency vs. average success rate. \vb TinyVLA and \cb OpenVLA.} Experiments on real-world Franka robot. The y-axis represents the average success rate across five real-world tasks, with the bubble diameter indicating the number of model parameters. Inference latency was measured on the same A6000 GPU for both models. Our results show that TinyVLA-H outperforms OpenVLA, achieving superior performance with 20 times less inference latency.}
    \label{fig:compare}
    \vspace{-0.7cm}
\end{figure}
\IEEEPARstart{T}{raining} multitasking robot imitators to operate in complex and uncertain environments faces considerable challenges due to limited data and the difficulty of learning physical motion~\cite{bharadhwaj2023roboagent,bridgedata}. Moreover, traditional robot models struggle to adapt to new scenes and tasks and are easily affected by distractors, lighting conditions, and background changes~\cite{diffusion-policy, ze20243d}. Modern methods typically leverage off-the-shelf Large Language Models (LLMs)~\cite{llama2, jiang2023mistral} for scene descriptions to generate object affordance, location, or heatmaps, followed by a predefined motion planner to complete the tasks~\cite{saycan, shi2023unleashing}. 

Recently, vision-language-action (VLA) models have garnered significant attention for their ability to extend pre-trained vision-language models to robotics using a next-token prediction approach. Notable works, such as RT-2~\cite{brohan2023rt-2} and OpenVLA~\cite{kim2024openvla}, have demonstrated impressive performance in multi-task learning and generalization. However, these methods suffer from a critical drawback: extremely slow inference speeds, largely due to their dependence on large vision-language models and auto-regressive action token generation. In robotics, inference speed is crucial for enabling robots to respond instantly to user queries, directly impacting user experience and the robot's overall effectiveness. In addition to the inference challenges, these models also require extensive pre-training on large-scale robotic datasets. For example, OpenVLA is pre-trained on the 970K-sample OpenX dataset~\cite{padalkar2023openx}, making the computational cost of training both expensive and resource-intensive. Given these challenges, a natural question arises:

\textbf{How can we build VLA models that retain the advantages of existing VLA models while being both fast and data-efficient?}

In this work, we propose \methodname, a compact vision-language-action model designed for fast inference. We identify two key factors in existing VLA models that contribute to their high inference latency: (1) they are built on large vision-language models, often exceeding 7 billion parameters, and (2) they generate discrete action tokens autoregressively, requiring repetitive inference for each degree of freedom. To overcome these challenges, we first train and employ a family of small yet powerful vision-language models with fewer than 1 billion parameters. Then, instead of using the next token prediction technique to predict action tokens independently, we attach a diffusion-based head to the pre-trained multimodal model for direct robot action output. Consequently, we find that this combination enables \methodname to retain the prior knowledge and generalization capabilities gained from vision-language data pre-training, even without training on large-scale robot datasets like OpenX~\cite{padalkar2023openx}. It efficiently adapts to new instruction and generalizes across various settings in a faster and more data-efficient manner.

In both simulations and real-world settings, our method demonstrates superior performance in multi-task learning compared to the baseline. For instance, in real-world experiments, \methodname-H achieves a 25.7\% higher success rate than OpenVLA, while using 5.5 times fewer parameters. In bimanual real-robot experiments, we find that OpenVLA, which heavily relies on OpenX robot data pretraining, struggles to perform in bimanual settings due to OpenX consisting only of single-arm data. In contrast, TinyVLA-H significantly outperforms OpenVLA in these tasks. Additionally, we observed that TinyVLA generalizes well across diverse settings, including observational and spatial generalization, often matching or even surpassing OpenVLA in certain cases. 
\\
Our contribution are the three folds:
\begin{itemize}
    \item We introduce a novel VLA architecture that combines lightweight vision-language models with a diffusion model, enabling fast inference, strong performance, and excellent generalization capabilities.
    \item We conducted extensive experiments in both simulated and real-world settings, encompassing single-arm and bimanual robot setups, to validate the effectiveness of our method.
    \item We demonstrate that strong VLA models can be trained without requiring large-scale robotic datasets, achieving both data-efficiency and high performance.
\end{itemize}

We believe that TinyVLA offers a novel perspective to building vision-language-action models for embodied control.
\section{Related Works}

\textbf{Vision-language models (VLMs).} VLMs connect vision and language and extend the reasoning ability of LLMs to process with multimodal input.  Numerous works have been proposed in this direction~\cite{minigpt4,llava,gemini,minigptv2}. These MLLMs typically have parameters ranging from 7B to 70B, making the inference cost-prohibitive and limiting the accessibility of MLLMs to a wider audience. Recently, a select number of studies~\cite{mobilevlmv2,zhu2024llava} have delved into the exploration of efficient multimodal, with a number of parameters less than 3B, from diverse angles. These models run efficiently, 

\textbf{Vision-language models for robot learning.} Robot learning~\cite{brohan2023rt-2, Zawalski24-ecot, wen2024object,fu2024mobile,black2023zero} is an crucial topic in the robotics. A number of works introduce vision-language models to the domain of robot learning, including using VLMs for high-level planning~\cite{sayplan}, task decomposition~\cite{saycan}, and formulate VLMs as a robot action predictor with end-to-end training~\cite{brohan2023rt-2, kim2024openvla, Zawalski24-ecot}. In this work, we explore two perspectives on using VLM as a robot action predictor, 1) how to use a more lightweight and fast VLM and 2) how to replace the autoregression model with a diffusion model.

\textbf{Multi-task robot learning.} Recent advances in multi-task robotic manipulation have yielded significant progress in executing complex tasks and generalizing to novel scenarios~\cite{brohan2022rt1, ze2024h, aldaco2024aloha}.  Leading methods often leverage extensive interaction data ~\cite{bcz} to train multi-task models. For example, RT-1~\cite{brohan2022rt1} underscores the benefits of task-agnostic training and RT-2~\cite{brohan2023rt-2} trains with mixed robot data and image-text pairs. PerAct~\cite{kumar2022pre} encodes language goals and shows its effectiveness in real robot experiments. Octo~\cite{team2024octo} uses cross-embodiment data for pertaining. This paper proposes a new approach to learning multi-task policy using a new form of vision-language-action models.

\vspace{-0.5cm}
\section{Method}

\begin{figure}[t]
    \centering
    \vspace{0.3cm}
    \includegraphics[width=0.48\textwidth]{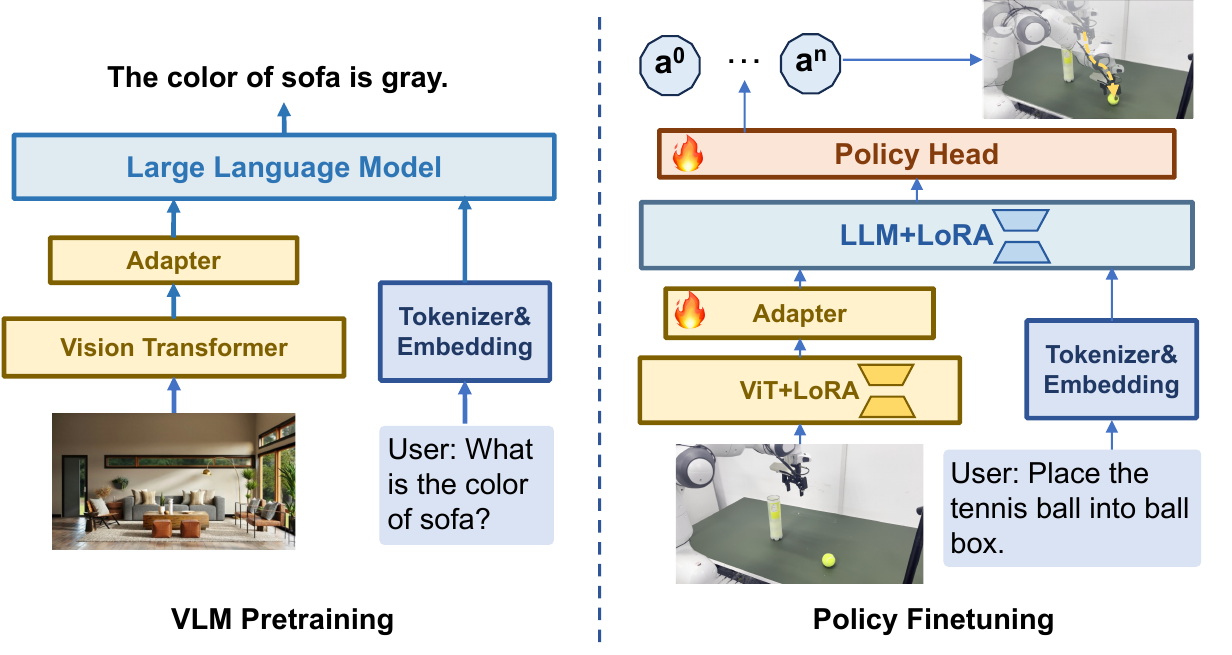}
    \caption{\textbf{Model architecture.} The left image illustrates the VLM pretraining pipeline, whereas the right image demonstrates the process of training \methodname~using robotic data. We adopt diffusion policy as our policy head.}\label{fig:framework}
    \vspace{-0.7cm}
\end{figure}

This section gives a comprehensive overview of our proposed TinyVLA. 
\methodname~encompasses several crucial designs: 1) We adopt a pre-trained VLM as the initialization of a policy network; 2) During training the robot data, we freeze the pre-trained parts and utilize the parameter-efficient fine-tuning technique LoRA~\cite{Lora}, where the trainable parameters account for only 5\% of the entire model; 3) We introduce a policy decoder that concatenated to pre-trained multimodal model through a simple but efficient linear projection and output the executable action of the robot. An illustration of \methodname is given in Figure~\ref{fig:framework}.

\subsection{Building TinyVLA with Efficient Vision-Language Models}
The initial step involves acquiring pre-trained vision-language models (VLM). While existing works typically focus on vision-language models with over three billion parameters, we trained a more compact vision-language model with parameters ranging from 70 million to 1.4 billion. Our model utilizes Pythia~\cite{pythia} as the language model backend. We then followed the training pipeline of LLaVA~\cite{llava}, using their vision-language dataset to train this family of VLMs. For robot data fine-tuning, we retained all modules from our VLM, including the visual backbone and the vision-language alignment module.

\subsection{Robot Data Finetuning for Manipulation}

\textbf{Frozen weights and low-rank adaptation.} We employ the parameter-efficient training method, LoRA~\cite{Lora}, which limits gradient updates to a low-dimensional space. This is achieved by modifying the weight matrix $W \in \mathbb{R}^{d \times k}$ to $W_0 + \Delta W = W_0 + BA$, with $B \in \mathbb{R}^{d \times r}$ and $A \in \mathbb{R}^{r \times k}$, where $r$ is significantly smaller than either $d$ or $k$. We incorporate low-rank matrices into the attention mechanisms' weights ($Q, K, V$) while freezing the remaining weights of the Transformer.

Furthermore, the model must preserve the intrinsic knowledge of the language models. The trainable parameters constitute only 5.0\% of the entire transformer's parameters. We posit that this approach enables the pre-trained model to process inputs with maximum linguistic fidelity while retaining flexibility. 
After training is completed, we apply re-parameterization techniques to integrate the LoRA module seamlessly into the standard language model, thereby enhancing inference speed. 

\textbf{Learning action with diffusion policy decoder.} We need a way to represent the action space to control the robot. One method is to use discrete tokenization for the actions, as has been done in RT-2. However, using tokenization for continuous or high-dimensional data has proven to be extremely challenging for training~\cite{lu2022unifiedio}, requires a huge amount of data~\cite{chen2022pix2seqv2, chen2023generalist}, and tends to converge to a single state~\cite{chen2021pix2seq}. Therefore, instead of converting actions into token space, we leverage the Diffusion Policy(DP)~\cite{diffusion-policy} as our policy head. DP formulates robot policies using Denoising Diffusion Probabilistic Models (DDPMs)~\cite{ddpms} which predicts the noise instead of direct actions. 

The whole framwork is illustrated in Figure~\ref{fig:framework}(right). And the pipeline can be splited into 3 steps. First, the visual-language model (VLM) backbone encodes raw observations and language instructions into multimodal embedding vectors. To handle the inherent variability in input sequence lengths, we employ an adaptive pooling layer followed by layer normalization, producing fixed and compact feature representations. Then, these normalized features are subsequently concatenated with the robot's proprioceptive state vector. The combined representation is processed through a 3-layer multilayer perceptron (MLP), generating conditional embeddings for the standard training process of DP. Finally, we utilize the standard training method of DP to train the whole VLA model. To preserve the intrinsic knowledge of the pretrained VLM, we implement different training strategies: The VLM is fine-tuned using low-rank adaptation(LoRA), while the DP head undergoes full-parameter training.


\section{Experiments}

In our experiments, we aim to study the following questions:
\begin{itemize}
    \item Does \methodname achieve a higher success rate in multitasking robotic manipulation compared to the baselines?
    \item Can \methodname interpret and follow novel instructions?
    \item Is \methodname capable of generalizing to unseen environments, adapting to new backgrounds, varying lighting conditions, changing camera view, and remaining robust against novel distractors?
    \item Does \methodname adhere to the scaling law, where a larger model size correlates with improved performance and better generalization?
\end{itemize}

\begin{figure}[t]
    \centering
    \vspace{0.3cm}
    \includegraphics[width=0.48\textwidth]{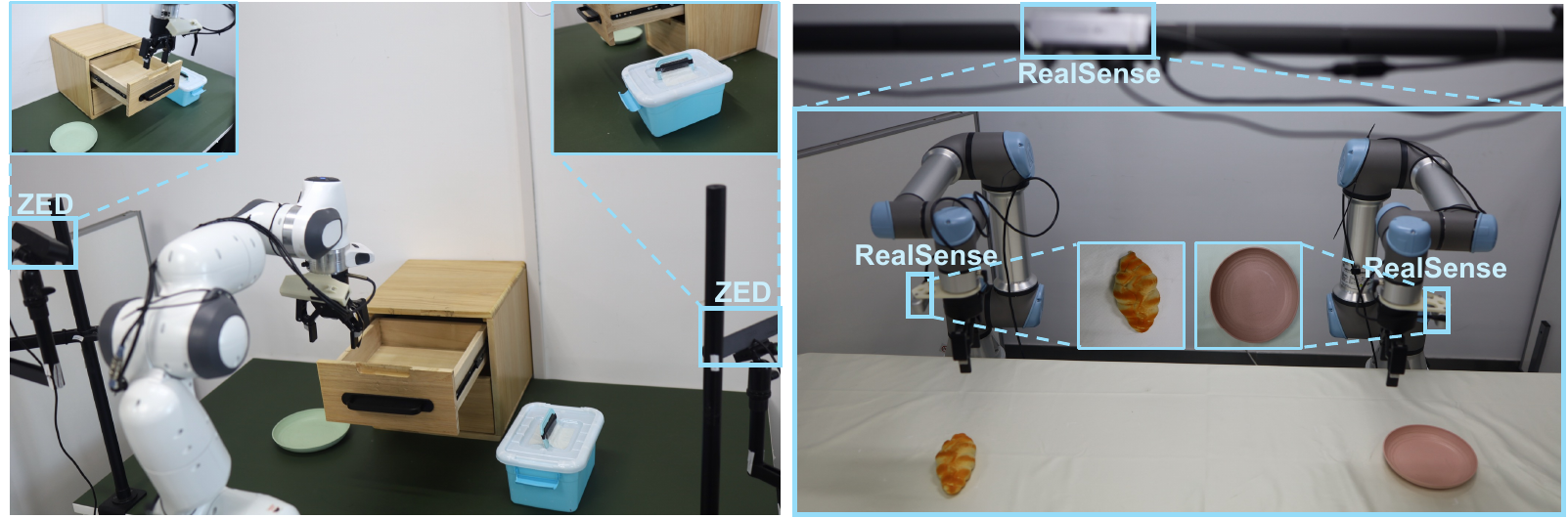}
    \caption{\textbf{Real robot settings.} The real robot setup for the single-arm Franka and bimanual UR5.}
    \label{fig:setting}
    \vspace{-0.7cm}
\end{figure}

\subsection{Experimental Setup}
To better distinguish the model sizes, we categorized \methodname~into three sizes based on the scale of the multimodal model: \methodname-S (Small), \methodname-B (Base) and \methodname-H (Huge).

\subsubsection{Simulation Benchmark}
We evaluate our approach on MetaWorld. The 50 tasks in MetaWorld~\cite{yu2020metaworld} can be categorized into multiple levels~\cite{seo2023masked}, i.e., easy, medium, hard, and very hard. 

\textbf{Baseline.} We compare our approach with the Diffusion Policy~\cite{diffusion-policy}. We report the average success rate. All methods are trained in a multi-task learning fashion with 50 demonstrations. It is evaluated with 3 seeds, and for each seed, the success rate was averaged over five different iterations.

\begin{table}[t]
\centering
\vspace{0.4cm}
\caption{Comparing \methodname~with Diffusion Policy in \textbf{simulation}. We report the average success rate on multiple tasks, We use \methodname-H as our method. \textbf{All methods are trained in a multi-task setting.}}
\label{tab:simulation}
\resizebox{0.48\textwidth}{!}{\begin{tabular}{c|cccc|c}
\toprule
 &  \multicolumn{5}{c}{Metaworld (50 tasks)}  \\
\textbf{Model $\setminus$ Tasks} &  Easy (28) & Medium (11)& Hard (6) & Very Hard (5) & Avg.\\
\midrule
Diffusion Policy~\cite{diffusion-policy}  & 23.1 & 10.7 & 1.9 & 6.1 & 10.5 \\ 
\textbf{\methodname-H} & \colorbox{AliceBlue}{\textbf{77.6}} & \colorbox{AliceBlue}{\textbf{21.5}} & \colorbox{AliceBlue}{\textbf{11.4}} & \colorbox{AliceBlue}{\textbf{15.8}} & \colorbox{AliceBlue}{\textbf{31.6}} \\ 
\bottomrule
\end{tabular}}
\vspace{-0.6cm}
\end{table}

\subsubsection{Real Robot Setup} \methodname~is both evaluated on a single arm setup utilizing a Franka Panda 7Dof robot arms and a bimanual setup with two UR5 robotic arms as illustrated in Figure ~\ref{fig:setting}. The single-arm scene is perceived via two external ZED 2 stereo cameras fixed on both sides of the robot. The bimanual robot's scene is captured by two cameras on wrists with an extra camera at the top. These cameras are Realsense D435i.

\textbf{Tasks.} In the single-arm setting, there are five tasks: 1) closing the drawer (CloseDrawer), 2) stacking the pink cube on top of the blue cube (StackCubes), 3) opening the lid of the box (OpenBox), 4) placing a tennis ball into the ball box, and 5) uprighting a tipped-over mug (FlipMug). In the bimanual robot experiment, we set up three tasks that involved cooperation between two arms: 1) transferring bread to a plate (TransferBread), 2) unzipping the bag and placing a tennis ball inside it (PlaceTennisBag), and 3)stacking cubes on a plate (StackCubes). It is worth noting that the action spaces of tasks vary considerably. For instance, \textit{flip mug} necessitates the robot to perform wide-ranging rotations to insert the gripper into the mug laterally, which is completely different from \textit{stack cubes} which is pick\&place type. The span of different trajectories within the same task varies markedly as well, e.g., the length of $stack~cubes$ trajectories ranges from 100 to 300. This provides \methodname with more challenges in learning to perform these tasks.

\textbf{Data collection.} We collect the dataset through teleoperation. We record the RGB stream from two camera views and robot states e.g., joint position during the whole robot control process. We record the robot gripper's width as a value between 0 and 1, where 0 represents fully closed and 1 represents fully open. \methodname predicts 7-dimensional actions, including position $(x,y,z)$, rotation $(roll, pitch, yaw)$ and $(gripper\_width)$. For all the tasks we do not add additional distractors except in the \textit{remove the lid of the box} task, in order to better evaluate the model's generalization capability to distractors. In total, we collected 100 trajectories for each task to balance data distribution across all 5 tasks. 

\textbf{Baseline.} We evaluated our method against Diffusion Policy (DP)~\cite{diffusion-policy}, Multimodal Diffusion~\cite{MDT2024} and OpenVLA~\cite{kim2024openvla}. We did a few modifications to ensure the comparison is fair. First of all, the vanilla OpenVLA is finetuned on a single view, which is incompatible with our approach. To ensure all camera views are utilized for OpenVLA, we process images from different views separately through the shared visual backbone, and then concatenate the visual tokens and feed them into the language models. Secondly, the vanilla DP does not incorporate language instructions. Therefore, following RT-1~\cite{brohan2022rt1} and YAY~\cite{shi2024yell}, we integrate language information into the visual backbone using FiLM~\cite{perez2018film}.

\begin{table*}[t]
\centering
\vspace{0.3cm}
\caption{\textbf{Quantitative results in real-world experiments.} We report the average success rate across multiple tasks and the count of trainable parameters for all models.}
\label{tab:main result}
\resizebox{0.95\textwidth}{!}{\begin{tabular}{c|c|cc|ccccc|c}
\toprule
 &  Pre-trained  & Total & Trainable  & \multicolumn{5}{c}{RealWorld(5 tasks)} &  \\
\textbf{Model $\setminus$ Tasks} & Trajectory & Params & Params & PlaceTennis & FlipMug & StackCubes & CloseDrawer& OpenBox & Avg.\\
\midrule
Diffusion Policy~\cite{diffusion-policy} & N/A & 111M &  $111M$ & 16.7$\pm$0.6& 30$\pm$0.2& 3.3$\pm$0.1&73.3$\pm$0.1& 53.3$\pm$0.1 & 35.3\\ 
Multimodal Diffusion\cite{MDT2024} & N/A & 230M &$230M$ & 23.3$\pm$0.3& 13.3$\pm$1.3& 6.7$\pm$0.3&36.7$\pm$0.3& 10.0$\pm$0 & 18.0\\ 
OpenVLA\cite{kim2024openvla} & 970K & 7.2B &$195M$ & 83.3$\pm$1.1& 51.7$\pm$3.1&40.0$\pm$0.1& 85.0$\pm$1& 81.7$\pm$0.6 & 68.3\\
\midrule
\methodname-S  & N/A & 422M &$101M$ & 8.3$\pm$0.1& 6.7$\pm$0.1& 6.7$\pm$0.1& 60.0$\pm$0.2& 35.0$\pm$0.3 & 23.3\\ 
\methodname-B  &N/A & 740M &$138M$ & 76.7$\pm$0.6 & 76.7$\pm$0.1& 71.7$\pm$0.1& 81.7$\pm$0.1& 80.0$\pm$0.2 & 77.4\\ 
\textbf{\methodname-H} & N/A & 1.3B &$143M$ &  \colorbox{AliceBlue}{\textbf{90.0$\pm$0.2}} & \colorbox{AliceBlue}{\textbf{98.3$\pm$0.1}} & \colorbox{AliceBlue}{\textbf{98.3$\pm$0.1}} & \colorbox{AliceBlue}{\textbf{96.7$\pm$0.3}} & \colorbox{AliceBlue}{\textbf{86.7$\pm$0.1}} & \colorbox{AliceBlue}{\textbf{94.0}}\\ 
\bottomrule
\end{tabular}}
\vspace{-0.4cm}
\end{table*}

\begin{table}[t]
\centering
\caption{\textbf{Quantitative results for bimanual UR5 real robot experiments.} We report the average success rate over 10 trials. All models are trained in multi-task settings.}
\vspace{-0.2cm}
\label{tab:bimanual result}

\resizebox{0.46\textwidth}{!}{\begin{tabular}{c|c|ccc}
\toprule
 & Trainable  & \multicolumn{3}{c}{RealWorld(3 tasks)}  \\
\textbf{Model $\setminus$ Tasks}& Params & PlaceBread & StackCubes & PlaceTennisBag \\
\midrule
DP~\cite{diffusion-policy} & $111M$ & 40.3$\pm$1.7& 31.3$\pm$1.3& 43$\pm$2.3\\ 
OpenVLA~\cite{kim2024openvla} & $195M$ & 0$\pm$0& 0$\pm$0& 0$\pm$0\\
\midrule
\textbf{\methodname-H} & $143M$ &  \colorbox{AliceBlue}{\textbf{76.7$\pm$2.3}} & \colorbox{AliceBlue}{\textbf{36.7$\pm$2.3}} & \colorbox{AliceBlue}{\textbf{30$\pm$1}} \\ 
\bottomrule
\end{tabular}}
\vspace{-0.7cm}
\end{table}

\begin{figure}[t]
    \vspace{0.3cm}
    \centering
    \includegraphics[width=0.45\textwidth]{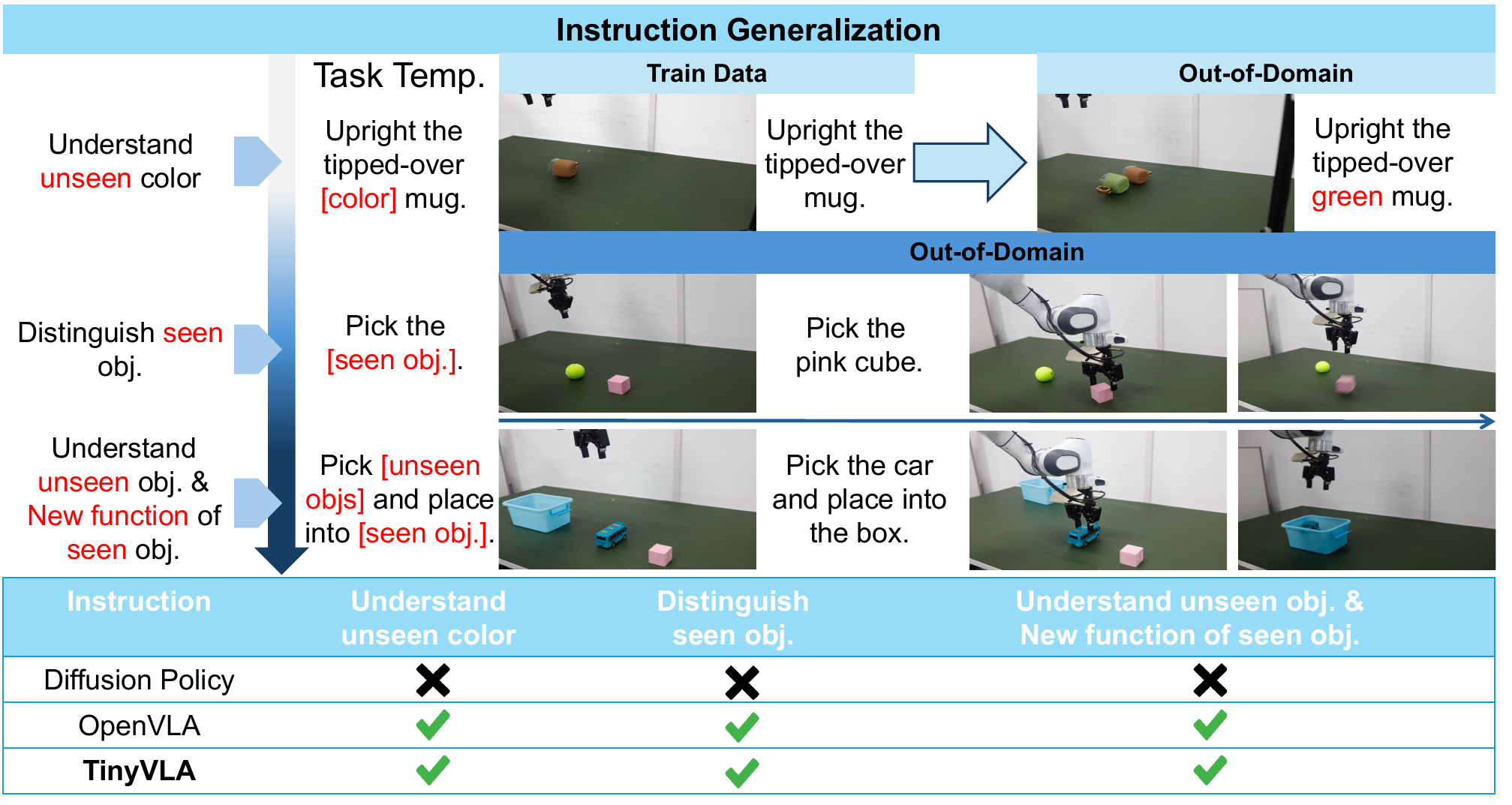}
    \caption{\textbf{Instruction Generalization.} We conducted three different types of instruction generalization experiments with progressively increasing difficulty.}\label{fig:instruct}
    \vspace{-0.8cm}

\end{figure}

\begin{figure*}[t]
    \centering
    \includegraphics[width=0.9\textwidth]{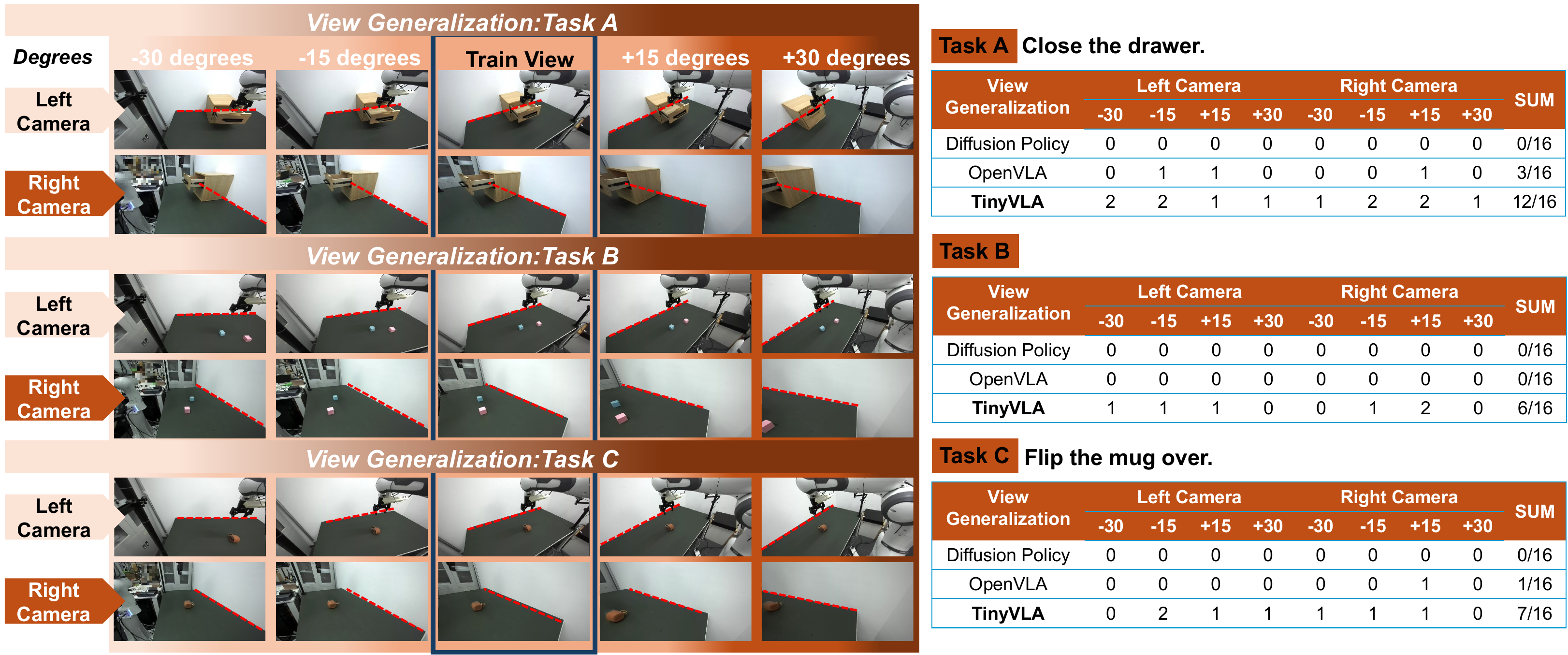}
        \caption{\textbf{View Generalization.} We evaluated the view generalization capability of our model in a new environment, which we designed to be as consistent with the training environment as possible. We tested 3 tasks respectively, and the results under 8 viewing angle changes (the two cameras each correspond to 4 changes).For each specified camera view(e.g. Left Camera -30 degrees), we evaluate each model for 2 trials.}
    \label{fig:view generalization}
    \vspace{-0.5cm}
\end{figure*}

\subsection{Experimental Results on Multi-Task Learning}

\textbf{Simulation experimental results.}
The experimental results are presented in Table~\ref{tab:simulation}. Specifically, \methodname's average success rate exceeds that of Diffusion Policy by 21.5\%. Notably, the performance disparity widens in more complex tasks; for instance, on the MetaWorld Hard scenario, \methodname's performance is sixfold better than that of Diffusion Policy. These results showcase the superiority of our proposed method.

\textbf{Real-world experimental results.}
The experimental results are shown in Table \ref{tab:main result}. We evaluate each model 20 trials per task in single arm setting. We report the mean and standard deviation of success rates across 3 checkpoints. Notably, \methodname-H attained a 98.3\% success rate in flipping a mug, stacking cubes, and a 90\% success rate in place tennis, leading a large margin over other baselines. 
Besides, the performance increases drastically from TinyVLA-S to TinyVLA-H which is adhere to scaling law. What's more, regarding the average success rate over five tasks, the result of TinyVLA-H surpasses OpenVLA by 25.7\%.

\subsection{Generalization to Unseen Instructions}

In this work, we investigate the generalization capabilities of \methodname-H, which demonstrates the best performance in both real-world scenarios and simulations. Since \methodname uses a pre-trained multimodal model as its backbone, we observe similar embodied capabilities driven by the rich world knowledge implicitly stored in these models, even though the fine-tuned version is not trained on question-answering pair data like RT-2~\cite{brohan2023rt-2}. As demonstrated in Figure~\ref{fig:instruct}, we evaluated with a fixed list of instructions (i.e., ``Pick the [object]"), where [object] are randomized objects that have not been seen in the training data. We use obj. as the abbreviation of objects in Figure~\ref{fig:instruct}. We test with three objects, a mug, a toy car, and a pink cube. 

The first level challenges \methodname to differentiate between an object with a seen color and one with an unseen color. Specifically, we placed two mugs of seen and unseen colors on the table and instructed \methodname to flip the green mug. Note that the green color has not been seen in the training data. \methodname successfully completed the task, demonstrating its inherent understanding of different object attributes.

The second level involves grasping the object. Both objects presented have been part of the training data. We asked the model to ``pick the cube". Despite the environment and instruction not being part of the training data, \methodname successfully picked up the cube. This indicates that \methodname effectively maps textual descriptions to physical objects.

To further increase the difficulty of the test, we designed the third level, where the model is instructed to ``pick a toy car" and ``place it into the box". The toy car is not in the training data. We placed a pink cube beside the toy car to assess whether the model could comprehend the instructions. Additionally, the command ``place into the box" introduces a new skill-object combination, suggesting that even though the object is familiar, its function has been altered. Successfully completing this task indicates that \methodname possesses the ability to recognize novel objects and identify new functionalities in familiar ones. 



\subsection{More Real-World Experiments: Bimanual Robot} We further conducted experiments on the Bimanual UR5 Robot, applying it to three distinct tasks: PlaceBread, StackCube, and PlaceTennisBag. These tasks vary significantly in both duration and required skills, posing challenges for training a multi-task policy model. As shown in Table~\ref{tab:bimanual result}, while the Diffusion Policy excels in the \textit{PlaceTennisBag} task, our \methodname-H model achieved an average success rate of 44.5\%, surpassing the Diffusion Policy's 38.2\%. Notably, the OpenVLA fails in every trial. We suspect this is because OpenVLA is pre-trained on the OpenX dataset, which consists entirely of single-arm robot data, making it ineffective when applied to bimanual robots.

\subsection{Experiments on Generalization}
In our approach, we integrate a pre-trained multimodal model with a Diffusion Policy head to generate robot actions. We demonstrate that leveraging a pre-trained multimodal model enhances the model's generalization capabilities across various perspectives. This integration not only optimizes action output but also significantly boosts the system's adaptability in diverse environments. For all experiments on generalization, we conduct one trial for each setting. Following DP3~\cite{ze20243d}, we use the same evaluation metrics. We use a cross mark to denote the failure of the model and a checkmark to indicate successful task completion.

\textbf{Generalization to new views.} Imitation learning, when trained on limited views, faces challenges in generalizing its learned capabilities to adapted views. In Figure~\ref{fig:view generalization}, we compare the view generalization capabilities of \methodname and Diffusion Policy. It appears that the Diffusion Policy is extremely sensitive to changes in viewpoint; even a slight shift can cause the model to fail. In contrast, \methodname demonstrates a certain degree of robustness in handling view generalization. For example, in tasks requiring high precision in object manipulation, such as Task B (StackCube) and Task C (FlipMug), our method can accommodate camera view shifts of up to 30 degrees to the left or right. Although it occasionally fails, \methodname still shows a significantly stronger view generalization compared to Diffusion Policy and OpenVLA, underscoring the benefits of using diffusion-based policy head.

\begin{figure}[t]
    \centering
    \vspace{0.3cm}
    \includegraphics[width=.4\textwidth]{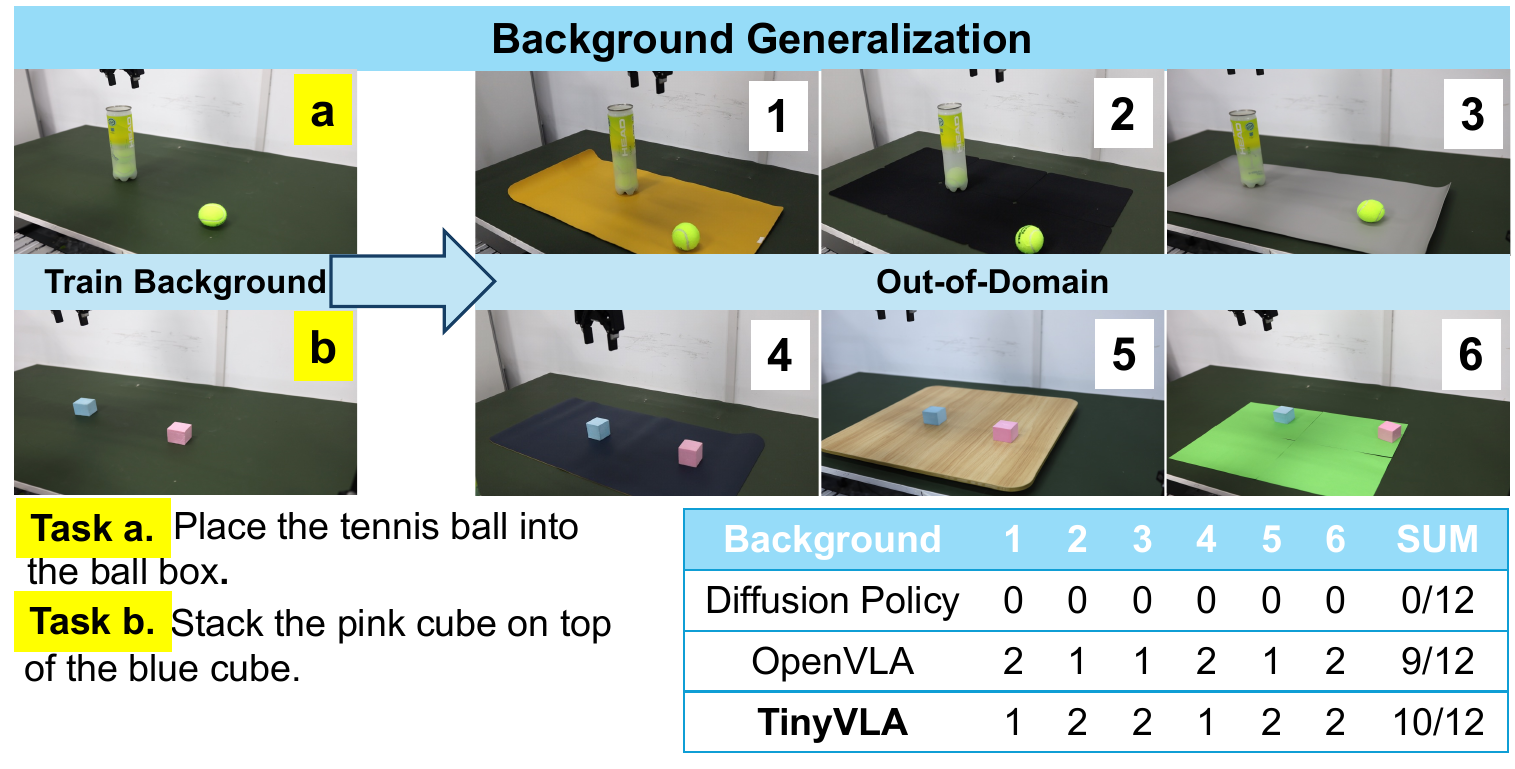}
    \caption{\textbf{Background Generalization.} We utilized six different backgrounds, testing three of them on Task a and the remaining three on Task b. For each background, we evaluate each model for two trials.}\label{fig:background generalization}
        \vspace{-0.7cm}
\end{figure}

\textbf{Background generalization:} We varied the background by using tablecloths of different colors and materials, including a wooden tabletop, mouse pad, desk mat, etc. In total, there are six distinct styles of backgrounds. We tested three of them on Task A and the remaining three on Task B. As shown in Figure~\ref{fig:background generalization}, our model accurately locates objects and successfully completes tasks across various scenarios, including position-sensitive tasks like placing a tennis ball, demonstrating performance comparable to the OpenVLA.

\begin{figure}[t]
    \centering
    \vspace{0.2cm}
    \includegraphics[width=0.48\textwidth]{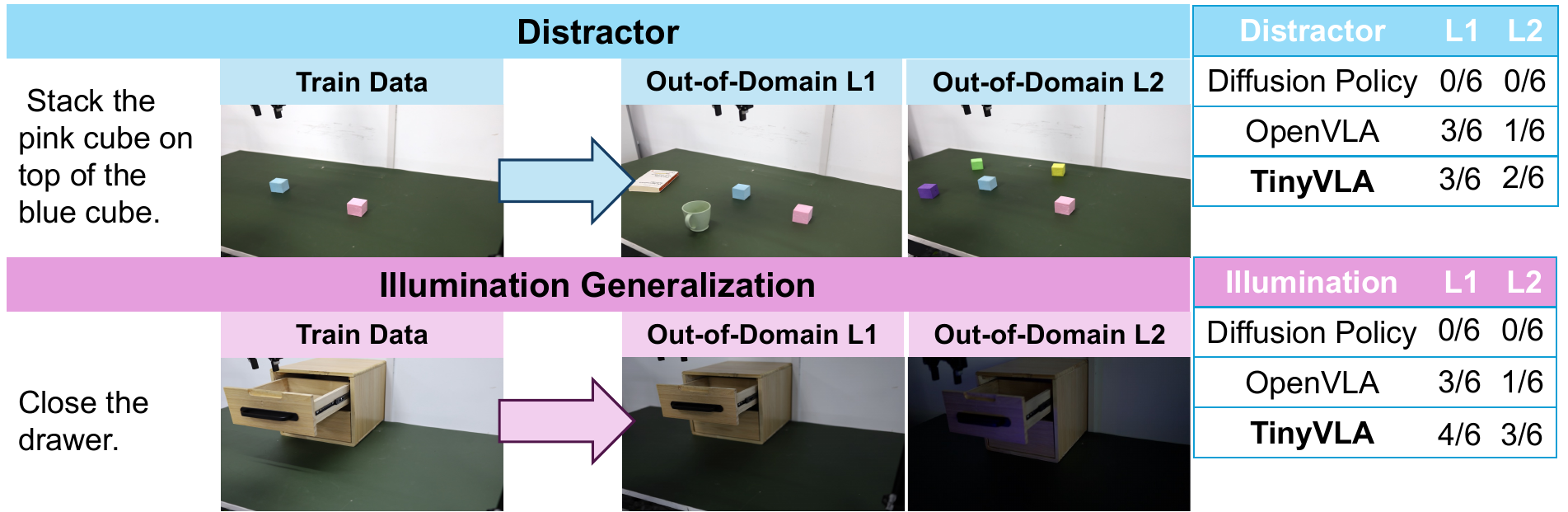}
    \caption{\textbf{Distractor \& Illumination Generalization.} For distractor settings, Level L1 involves the addition of objects such as books and cups, which are unrelated to the task. Level L2 involves the inclusion of identical cubes in various colors, adding complexity to the visual environment. For illumination settings, Level L1 represents conditions with reduced lighting, while Level L2 describes scenarios with minimal lighting. For each specified setting (i.e., Distractor L1), we evaluate each model for six trials.}\label{fig:distractor_genralization}
       \vspace{-0.7cm}
\end{figure}

\textbf{Generalization to different light conditions:} Regarding light conditions, conventional policy networks are sensitive to variations in lighting. As shown in Figure~\ref{fig:distractor_genralization}(bottom), we analyze the impact of three different lighting scenarios. The left image represents our training data. The middle image depicts the scenario when the overhead lights are turned off, and the right image shows conditions with all our lights turned off. We observe that \methodname remains unaffected by these variations in lighting, whereas the OpenVLA fails to complete the task under low light conditions. This reinforces our previous findings, confirming that our method is highly robust against changes in background lighting.

\textbf{Generalization to distractor:} It is known that the diffusion policy is sensitive to distractors, meaning that when objects not present in the collected data appear, the policy typically fails to complete the tasks. Indeed, adding strong augmentation could alleviate this problem. We aim to study whether the model, without data augmentation, could be robust to the appearance of distractors. In Figure~\ref{fig:distractor_genralization}~(top), we present the StackCube task featuring an additional distractor, categorized into two difficulty levels. Our model effectively manages both types of distractors at each difficulty level, whereas the Diffusion Policy and OpenVLA struggles with both. This demonstrates that utilizing a pre-trained multimodal model significantly enhances generalization capabilities in the presence of distractors.

\begin{figure}[t]
    \vspace{0.3cm}
    \centering
    \includegraphics[width=.48\textwidth]{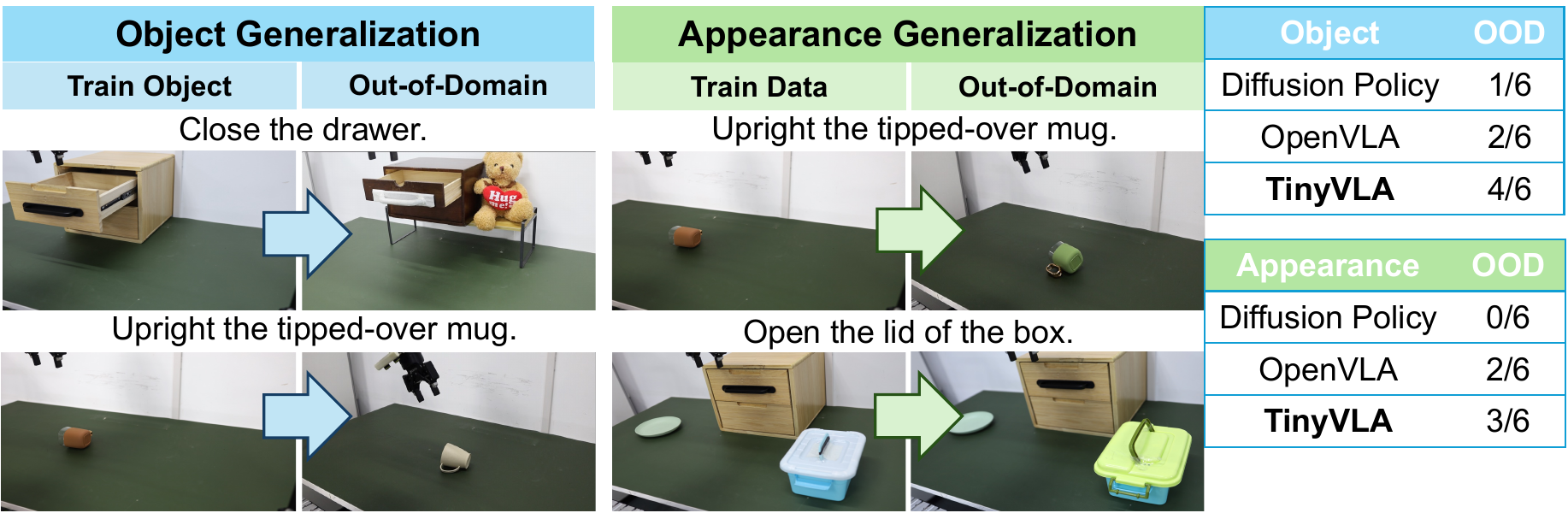}
    \caption{\textbf{Object \& Appearance generalization.} For object generalization, we replace the objects with previously unseen ones that have different shapes or colors. For appearance generalization, we only alter the colors of the objects. For each specified setting, we evaluate each model for 6 trials.}\label{fig:object generalization}
    \vspace{-0.5cm}
\end{figure}

\begin{figure}[t]
    \centering
    \includegraphics[width=.48\textwidth]{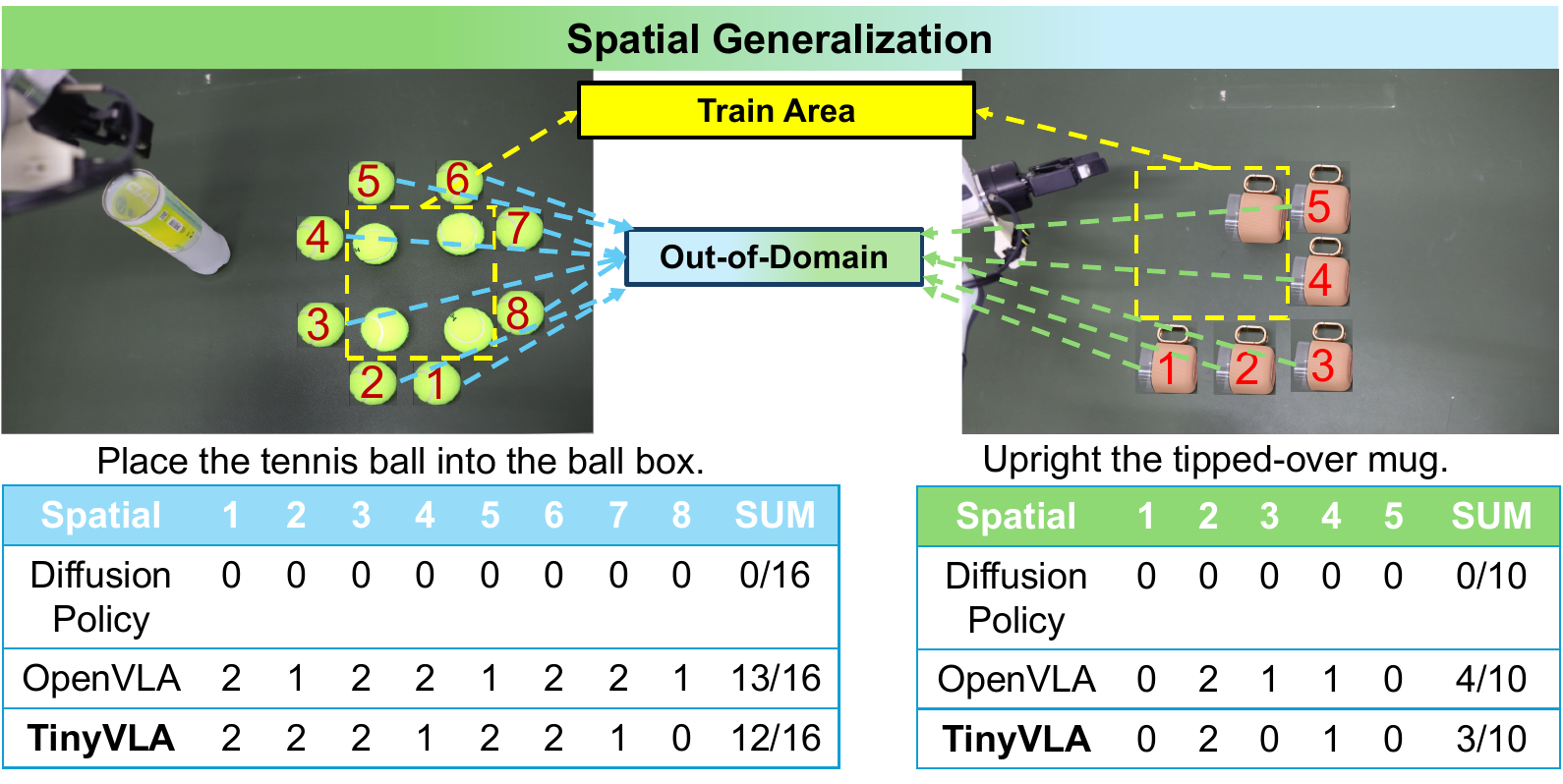}
    \caption{\textbf{Spatial generalization.} We conducted evaluations at multiple positions thoroughly outside the training zone on two position-sensitive tasks:place tennis and flip mug. For each out-of-distribution positions, we evaluate each model for 2 trials.}\label{fig:spation generalization}
    \vspace{-0.7cm}
\end{figure}

\vspace{-0.5cm}
\subsection{Spatial Generalization}
Spatial generalization~\cite{doumas2022theory, toyer2020magical, yin2023spatial, yarats2020image} refers to the generalization to
unseen setup of objects (entities) locations in one task, which instead requires physical common sense about space and object. In Figure~\ref{fig:spation generalization}, we present the spatial generalization performance of our methods. Intriguingly, although our \methodname~model was not trained on the specific locations of objects in the training dataset, it successfully completes tasks involving these objects. Furthermore, we have tested our method in locations significantly distant from those in our training data, as illustrated in Figure~\ref{fig:spation generalization}. We observe that OpenVLA performs slightly better than our approach, likely because it is trained on large-scale robotic data, allowing the model to "see" more diverse robot actions during pre-training. In contrast, the Diffusion Policy, which is trained on the same data as our model, consistently fails to generalize spatially across the tested locations.

\subsection{Visual Generalization} Visual generalization pertains to the adaptation to novel visual textures. In robotic manipulation tasks, this type of generalization can be seen in variations in background color, object texture, or ambient lighting. These visual changes do not impact the fundamental task structure, such as the positioning of objects and targets. Instead, they necessitate that the robot accurately interpret the semantic meanings associated with these visual cues. 

\textbf{Appearance generalization:} We altered the color of the target objects, as demonstrated in Figure~\ref{fig:object generalization} (right). Initially, the mug was brown, and the lid was white; we then modified their colors. We observe that \methodname~successfully generalizes to objects with varying colors, demonstrating a capability similar to that of OpenVLA. Notably, our approach achieves appearance generalization without relying on data augmentation during training. This indicates that the generalization capability of our model stems from the pre-trained vision-language data. 

\begin{figure}[h]
        \vspace{0.3cm}
    \centering
    \includegraphics[width=.35\textwidth]{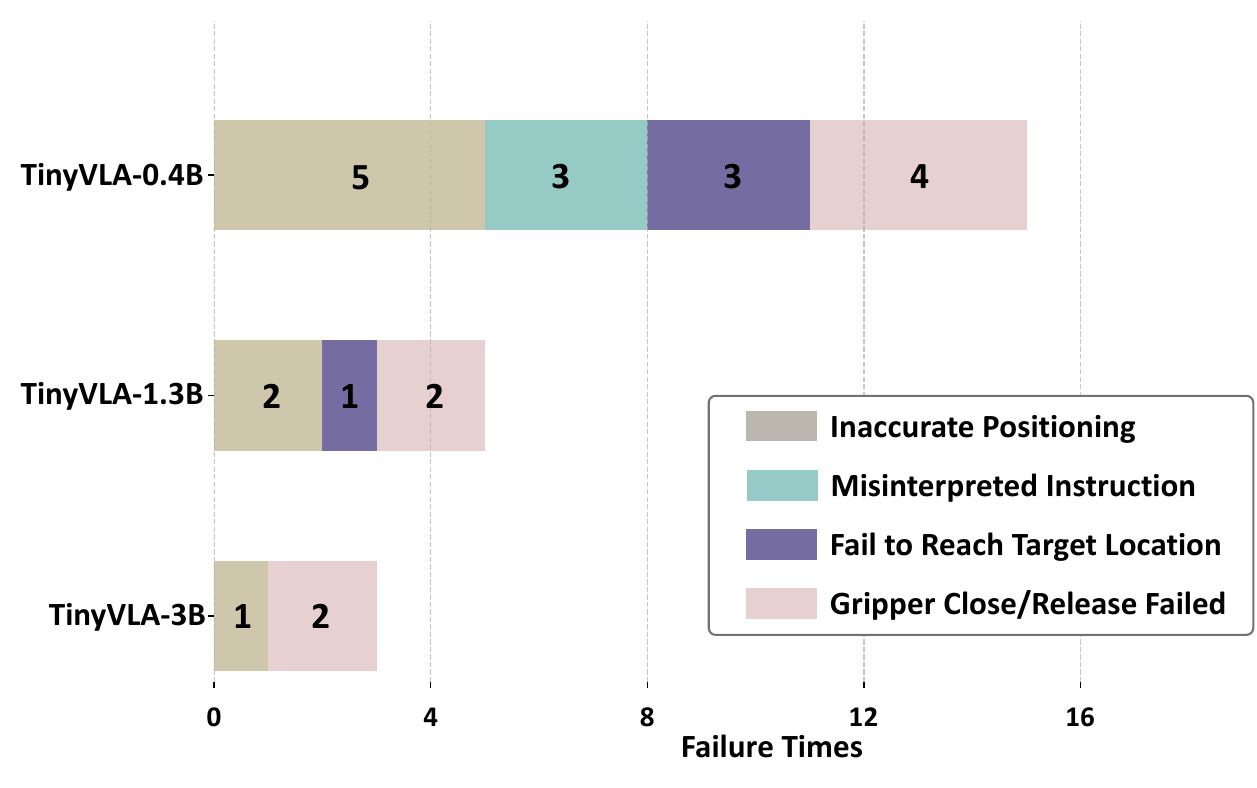}
    \caption{Types of failure for TinyVLA with different sizes of pre-trained vision-language models.}\label{fig:fail_case_study}
    \vspace{-0.7cm}
\end{figure}

\section{Ablation Study}
\subsection{Trade-off between size of VLM and TinyVLA's Performance}
Our main experiments (Table~\ref{tab:main result} and Table~\ref{tab:bimanual result}) demonstrate that our method adheres to the scaling law: as model size increases, the average success rate across tasks improves accordingly. To understand the underlying reasons for this trade-off between model size and performance, we conducted a failure case analysis.  We evaluated three TinyVLA variants: TinyVLA-0.4B, TinyVLA-1.3B, and TinyVLA-3B. The first two were used in the main experiments, while TinyVLA-3B utilizes the pre-trained PaliGemma model~\cite{beyer2024paligemma}. Each model was tested on one of four tasks: PlaceTennis, FlipMug, StackCubes on a Franka robot, and PlaceTennisBag on a bimanual robot. Each task was evaluated six times, and the total number of failures was recorded. The results are presented in Figure~\ref{fig:fail_case_study}).

Our analysis reveals that VLM size significantly impacts task success. For example, TinyVLA-0.4B failed three times due to misinterpreting instructions, likely because the smaller VLM has limited language comprehension capabilities. This issue was resolved when we increased the model size to 1.3B.  Furthermore, increasing the model size mitigated failures related to inaccurate positioning and reaching incorrect target locations. This improvement can be attributed to the use of models like PaliGemma, which are trained on localization data and possess richer visual feature representations, leading to enhanced localization abilities.

\subsection{Choice of Policy Model}
Our TinyVLA model exhibits strong performance and generalization capabilities. This success is largely attributed to the integration of a pre-trained Vision-Language Model (VLM) with a diffusion model.  However, this raises a crucial question: how essential is the diffusion model to this architecture? Could alternative methods achieve comparable results? To investigate this, we compared TinyVLA's performance with two different policy networks: a vanilla multi-layer perceptron (MLP) commonly used for behavior cloning, and an action chunking transformer (ACT)~\cite{zhaoaloha} known for generating stable and smooth actions, as shown in Table~\ref{tab:difference_head}. Our findings demonstrate that the diffusion model significantly outperforms both ACT and MLP. Notably, the MLP-based approach failed entirely across all tasks, likely due to the limited capacity of the MLP layers compared to the VLM, hindering effective optimization. While ACT demonstrated some success, our diffusion model achieved a considerably higher average success rate across all tasks. These results underscore the significant advantage of employing a diffusion model over alternative policy networks in our TinyVLA framework.

\subsection{Which Part of TinyVLA makes it Fast?}
A key advantage of \methodname~is its lightweight design and superior speed compared to OpenVLA. As demonstrated in Figure~\ref{fig:compare}, our largest model, \methodname-H, achieves a higher average success rate than OpenVLA while utilizing 5.5 times fewer parameters and operating 20 times faster. This section analyzes the primary contributor to this significant speed advantage. Specifically, we replaced the Prismatic-7B VLM backbone in OpenVLA with the same architecture in \methodname. We demonstrate the comparison in Table~\ref{tab:new_inf_cost}. We observed a reduction in per-action prediction time from 292ms to 140ms. Despite this 2x speed increase, OpenVLA remains 10 times slower than TinyVLA-H, with a similar number of parameters. This result highlights that the speed of TinyVLA stems not only from utilizing a smaller VLM but also from employing a diffusion model for action prediction. This approach avoids the computationally expensive autoregressive generation of action tokens, improving test-time speed.

\begin{table}[h]
\centering
\vspace{0.2cm}
\caption{Inference latency is measured quantitatively and reported in milliseconds (ms). The reported values represent the time required for a single action prediction by the OpenVLA-1B/7B and TinyVLA-1B models. Experiments are conducted using a single A6000 GPU.}
\label{tab:new_inf_cost}
\resizebox{0.4\textwidth}{!}{\begin{tabular}{c|c}
\toprule
 \multicolumn{2}{c}{Inference Latency on A6000 GPU}  \\
 \midrule
 OpenVLA-7B $\rightarrow$ OpenVLA-1B & TinyVLA-1B  \\
\midrule
 292 $ms$ $\rightarrow$ 140 $ms$ & 14 $ms$ \\ 
\bottomrule
\end{tabular}}
\vspace{-0.5cm}
\end{table}

\begin{table}[h]
\centering
\caption{Ablation study on different choices of policy model. We choose \methodname-H as our base model and replace the diffusion model with ACT model~\cite{zhaoaloha} and vanilla MLP head. We report the success rate on 5 real robot tasks.}
\label{tab:difference_head}
\resizebox{0.48\textwidth}{!}{\begin{tabular}{c|c|c|c|c|c}
\toprule
Policy Head & PlaceTennis & FlipMug & StackCubes & CloseDrawer & OpenBox \\
\midrule
Multi-Layer Perceptron & 0 $\pm$ 0 & 0 $\pm$ 0 & 0 $\pm$ 0 & 0 $\pm$ 0 & 0 $\pm$ 0 \\ 
Action Chunking Transformer & 13.3$\pm$0.1 & 8.3$\pm$0.1 & 8.3$\pm$0.3 & 13.3$\pm$0.1 & 23.3$\pm$0.1 \\
Diffusion Model & \textbf{90$\pm$0.2} & \textbf{98.3$\pm$0.1} & \textbf{98.3$\pm$0.1} & \textbf{96.7$\pm$0.3} & \textbf{86.7$\pm$0.1} \\
\bottomrule
\end{tabular}}
\vspace{-0.5cm}
\end{table}

\section{Conclusion}
\label{sec:conclusion}
In this work, we explore the potential of leveraging pre-trained multimodal models for robotic manipulation. Our approach overcomes the limitations of previous methods by enabling fast inference and significantly reducing the computational resources required for training. We demonstrate the effectiveness of our method through both simulation and real-world experiments. We believe our approach offers a novel solution for building fast, data-efficient VLA models.

\section*{Acknowledgement}
We sincerely thank Yanjie Ze for contributions in discussions and paper review. We also thank Yong Wang for his assistance with the hardware setup.


\bibliographystyle{IEEEtran}
\bibliography{mai.bbl}

\begin{thebibliography}{10}
\providecommand{\url}[1]{#1}
\csname url@samestyle\endcsname
\providecommand{\newblock}{\relax}
\providecommand{\bibinfo}[2]{#2}
\providecommand{\BIBentrySTDinterwordspacing}{\spaceskip=0pt\relax}
\providecommand{\BIBentryALTinterwordstretchfactor}{4}
\providecommand{\BIBentryALTinterwordspacing}{\spaceskip=\fontdimen2\font plus
\BIBentryALTinterwordstretchfactor\fontdimen3\font minus \fontdimen4\font\relax}
\providecommand{\BIBforeignlanguage}[2]{{%
\expandafter\ifx\csname l@#1\endcsname\relax
\typeout{** WARNING: IEEEtran.bst: No hyphenation pattern has been}%
\typeout{** loaded for the language `#1'. Using the pattern for}%
\typeout{** the default language instead.}%
\else
\language=\csname l@#1\endcsname
\fi
#2}}
\providecommand{\BIBdecl}{\relax}
\BIBdecl

\bibitem{bharadhwaj2023roboagent}
H.~Bharadhwaj, J.~Vakil \emph{et~al.}, ``Roboagent: Generalization and efficiency in robot manipulation via semantic augmentations and action chunking,'' in \emph{ICRA 2024}.\hskip 1em plus 0.5em minus 0.4em\relax IEEE, 2024, pp. 4788--4795.

\bibitem{bridgedata}
F.~Ebert, Y.~Yang \emph{et~al.}, ``Bridge data: Boosting generalization of robotic skills with cross-domain datasets,'' \emph{RSS}, 2022.

\bibitem{diffusion-policy}
C.~Chi, S.~Feng \emph{et~al.}, ``Diffusion policy: Visuomotor policy learning via action diffusion,'' \emph{RSS}, 2023.

\bibitem{ze20243d}
Y.~Ze, G.~Zhang \emph{et~al.}, ``3d diffusion policy: Generalizable visuomotor policy learning via simple 3d representations,'' in \emph{Proceedings of Robotics: Science and Systems (RSS)}, 2024.

\bibitem{llama2}
H.~Touvron, L.~Martin \emph{et~al.}, ``Llama 2: Open foundation and fine-tuned chat models,'' \emph{arXiv preprint arXiv:2307.09288}, 2023.

\bibitem{jiang2023mistral}
A.~Q. Jiang, A.~Sablayrolles, A.~Mensch, C.~Bamford, D.~S. Chaplot, D.~d.~l. Casas, F.~Bressand, G.~Lengyel, G.~Lample, L.~Saulnier \emph{et~al.}, ``Mistral 7b,'' \emph{arXiv preprint arXiv:2310.06825}, 2023.

\bibitem{saycan}
M.~Ahn, A.~Brohan \emph{et~al.}, ``Do as i can, not as i say: Grounding language in robotic affordances,'' \emph{arXiv preprint arXiv:2204.01691}, 2022.

\bibitem{shi2023unleashing}
R.~Shi, Y.~Liu, Y.~Ze, S.~S. Du, and H.~Xu, ``Unleashing the power of pre-trained language models for offline reinforcement learning,'' \emph{arXiv preprint arXiv:2310.20587}, 2023.

\bibitem{brohan2023rt-2}
A.~Brohan, N.~Brown \emph{et~al.}, ``Rt-2: Vision-language-action models transfer web knowledge to robotic control,'' \emph{arXiv preprint arXiv:2307.15818}, 2023.

\bibitem{kim2024openvla}
M.~J. Kim, K.~Pertsch \emph{et~al.}, ``Openvla: An open-source vision-language-action model,'' \emph{8th Annual Conference on Robot Learning}, 2024.

\bibitem{padalkar2023openx}
A.~Padalkar, A.~Pooley \emph{et~al.}, ``Open x-embodiment: Robotic learning datasets and rt-x models,'' \emph{arXiv preprint arXiv:2310.08864}, 2023.

\bibitem{minigpt4}
D.~Zhu, J.~Chen \emph{et~al.}, ``Mini{GPT}-4: Enhancing vision-language understanding with advanced large language models,'' in \emph{The Twelfth International Conference on Learning Representations}, 2024.

\bibitem{llava}
H.~Liu, C.~Li, Q.~Wu, and Y.~J. Lee, ``Visual instruction tuning,'' in \emph{Thirty-seventh Conference on Neural Information Processing Systems}, 2023.

\bibitem{gemini}
G.~Team, R.~Anil \emph{et~al.}, ``Gemini: a family of highly capable multimodal models,'' \emph{arXiv preprint arXiv:2312.11805}, 2023.

\bibitem{minigptv2}
J.~Chen, D.~Zhu \emph{et~al.}, ``Minigpt-v2: large language model as a unified interface for vision-language multi-task learning,'' \emph{arXiv preprint arXiv:2310.09478}, 2023.

\bibitem{mobilevlmv2}
X.~Chu, L.~Qiao \emph{et~al.}, ``Mobilevlm v2: Faster and stronger baseline for vision language model,'' \emph{arXiv preprint arXiv:2402.03766}, 2024.

\bibitem{zhu2024llava}
Y.~Zhu, M.~Zhu, N.~Liu, Z.~Ou, X.~Mou, and J.~Tang, ``Llava-phi: Efficient multi-modal assistant with small language model,'' \emph{arXiv preprint arXiv:2401.02330}, 2024.

\bibitem{Zawalski24-ecot}
M.~Zawalski, W.~Chen, K.~Pertsch, O.~Mees, C.~Finn, and S.~Levine, ``Robotic control via embodied chain-of-thought reasoning,'' in \emph{Conference on Robot Learning (CoRL)}, vol. 270, 2024, pp. 3157--3181.

\bibitem{wen2024object}
J.~Wen, Y.~Zhu, M.~Zhu, J.~Li, Z.~Xu, Z.~Che, C.~Shen, Y.~Peng, D.~Liu, F.~Feng \emph{et~al.}, ``Object-centric instruction augmentation for robotic manipulation,'' \emph{arXiv preprint arXiv:2401.02814}, 2024.

\bibitem{fu2024mobile}
Z.~Fu, T.~Z. Zhao, and C.~Finn, ``Mobile aloha: Learning bimanual mobile manipulation with low-cost whole-body teleoperation,'' \emph{arXiv preprint arXiv:2401.02117}, 2024.

\bibitem{black2023zero}
K.~Black, M.~Nakamoto, P.~Atreya, H.~Walke, C.~Finn, A.~Kumar, and S.~Levine, ``Zero-shot robotic manipulation with pretrained image-editing diffusion models,'' \emph{arXiv preprint arXiv:2310.10639}, 2023.

\bibitem{sayplan}
K.~Rana, J.~Haviland \emph{et~al.}, ``Sayplan: Grounding large language models using 3d scene graphs for scalable task planning,'' in \emph{7th Annual Conference on Robot Learning}, 2023.

\bibitem{brohan2022rt1}
A.~Brohan, N.~Brown \emph{et~al.}, ``Rt-1: Robotics transformer for real-world control at scale,'' \emph{arXiv preprint arXiv:2212.06817}, 2022.

\bibitem{ze2024h}
Y.~Ze, Y.~Liu \emph{et~al.}, ``H-index: Visual reinforcement learning with hand-informed representations for dexterous manipulation,'' \emph{Advances in Neural Information Processing Systems}, vol.~36, 2024.

\bibitem{aldaco2024aloha}
J.~Aldaco, T.~Armstrong \emph{et~al.}, ``Aloha 2: An enhanced low-cost hardware for bimanual teleoperation,'' \emph{arXiv preprint arXiv:2405.02292}, 2024.

\bibitem{bcz}
E.~Jang, A.~Irpan \emph{et~al.}, ``Bc-z: Zero-shot task generalization with robotic imitation learning,'' in \emph{Conference on Robot Learning}.\hskip 1em plus 0.5em minus 0.4em\relax PMLR, 2022, pp. 991--1002.

\bibitem{kumar2022pre}
A.~Kumar, A.~Singh, F.~Ebert, Y.~Yang, C.~Finn, and S.~Levine, ``Pre-training for robots: Offline rl enables learning new tasks from a handful of trials,'' \emph{arXiv preprint arXiv:2210.05178}, 2022.

\bibitem{team2024octo}
{Octo Model Team}, D.~Ghosh, H.~Walke \emph{et~al.}, ``Octo: An open-source generalist robot policy,'' in \emph{Proceedings of Robotics: Science and Systems}, Delft, Netherlands, 2024.

\bibitem{Lora}
E.~J. Hu, yelong shen \emph{et~al.}, ``Lo{RA}: Low-rank adaptation of large language models,'' in \emph{International Conference on Learning Representations}, 2022.

\bibitem{pythia}
S.~Biderman, H.~Schoelkopf \emph{et~al.}, ``Pythia: A suite for analyzing large language models across training and scaling,'' in \emph{International Conference on Machine Learning}.\hskip 1em plus 0.5em minus 0.4em\relax PMLR, 2023, pp. 2397--2430.

\bibitem{lu2022unifiedio}
J.~Lu, C.~Clark, R.~Zellers, R.~Mottaghi, and A.~Kembhavi, ``Unified-io: A unified model for vision, language, and multi-modal tasks,'' in \emph{The Eleventh International Conference on Learning Representations}, 2022.

\bibitem{chen2022pix2seqv2}
T.~Chen, S.~Saxena, L.~Li, T.-Y. Lin, D.~J. Fleet, and G.~E. Hinton, ``A unified sequence interface for vision tasks,'' \emph{Advances in Neural Information Processing Systems}, vol.~35, pp. 31\,333--31\,346, 2022.

\bibitem{chen2023generalist}
T.~Chen, L.~Li \emph{et~al.}, ``A generalist framework for panoptic segmentation of images and videos,'' in \emph{Proceedings of the IEEE/CVF International Conference on Computer Vision}, 2023, pp. 909--919.

\bibitem{chen2021pix2seq}
T.~Chen, S.~Saxena, L.~Li, D.~J. Fleet, and G.~Hinton, ``Pix2seq: A language modeling framework for object detection,'' \emph{arXiv preprint arXiv:2109.10852}, 2021.

\bibitem{ddpms}
J.~Ho, A.~Jain, and P.~Abbeel, ``Denoising diffusion probabilistic models,'' \emph{Advances in neural information processing systems}, vol.~33, pp. 6840--6851, 2020.

\bibitem{yu2020metaworld}
T.~Yu, D.~Quillen \emph{et~al.}, ``Meta-world: A benchmark and evaluation for multi-task and meta reinforcement learning,'' in \emph{Conference on robot learning}.\hskip 1em plus 0.5em minus 0.4em\relax PMLR, 2020, pp. 1094--1100.

\bibitem{seo2023masked}
Y.~Seo, D.~Hafner, H.~Liu, F.~Liu, S.~James, K.~Lee, and P.~Abbeel, ``Masked world models for visual control,'' in \emph{Conference on Robot Learning}.\hskip 1em plus 0.5em minus 0.4em\relax PMLR, 2023, pp. 1332--1344.

\bibitem{MDT2024}
M.~Reuss, {\"O}.~E. Ya{\u{g}}murlu, F.~Wenzel, and R.~Lioutikov, ``Multimodal diffusion transformer: Learning versatile behavior from multimodal goals,'' \emph{Robotics: Science and Systems}, 2024.

\bibitem{shi2024yell}
L.~X. Shi, Z.~Hu \emph{et~al.}, ``Yell at your robot: Improving on-the-fly from language corrections,'' \emph{arXiv preprint arXiv:2403.12910}, 2024.

\bibitem{perez2018film}
E.~Perez, F.~Strub \emph{et~al.}, ``Film: Visual reasoning with a general conditioning layer,'' in \emph{Proceedings of the AAAI conference on artificial intelligence}, vol.~32, no.~1, 2018.

\bibitem{doumas2022theory}
L.~A. Doumas, G.~Puebla, A.~E. Martin, and J.~E. Hummel, ``A theory of relation learning and cross-domain generalization.'' \emph{Psychological review}, vol. 129, no.~5, p. 999, 2022.

\bibitem{toyer2020magical}
S.~Toyer, R.~Shah, A.~Critch, and S.~Russell, ``The magical benchmark for robust imitation,'' \emph{Advances in Neural Information Processing Systems}, vol.~33, pp. 18\,284--18\,295, 2020.

\bibitem{yin2023spatial}
Z.-H. Yin, Y.~Gao, and Q.~Chen, ``Spatial generalization of visual imitation learning with position-invariant regularization,'' in \emph{RSS 2023 Workshop on Symmetries in Robot Learning}, 2023.

\bibitem{yarats2020image}
D.~Yarats, I.~Kostrikov, and R.~Fergus, ``Image augmentation is all you need: Regularizing deep reinforcement learning from pixels,'' in \emph{International conference on learning representations}, 2020.

\bibitem{beyer2024paligemma}
L.~Beyer, A.~Steiner \emph{et~al.}, ``Paligemma: A versatile 3b vlm for transfer,'' \emph{arXiv preprint arXiv:2407.07726}, 2024.

\bibitem{zhaoaloha}
T.~Z. Zhao, J.~Tompson \emph{et~al.}, ``Aloha unleashed: A simple recipe for robot dexterity,'' in \emph{8th Annual Conference on Robot Learning}, 2024.

\end{thebibliography}

%




\end{document}